# A Machine-Learning-Based Gas Lift Optimization Workflow for Unconventional Fields

Sha(Sasha) Miao*[1], Alexandra Vendetti[2], Logan Smart[2], Gunta Chomchalerm[1], Yang Chen[1], Christopher Frazier[2], Dustin Haralson[2], Jeremy Sorenson[2], Xiao Ma[1], Huafei Sun[1], Aaron Shinn[2], Haining Zheng[1], Xiao-Hui Wu[1], Peng Xu[1], 1. ExxonMobil Technology and Engineering Company, 2. ExxonMobil Upstream Company

## Abstract

In this paper, we present an automated data-driven workflow using Machine Learning (ML) for gas lift optimization in unconventional fields. This workflow integrates a ML model that accurately forecasts the Gas Lift Performance Curve, and a Bayesian Optimization Framework to solve for the optimal gas injection rates under the constraints of facility capacity. The ML model leverages the historical production time series data without requiring downhole gauges or multi-rate well tests. We piloted this workflow on 30 wells across 5 well pads in Bakken and obtained >5% production uplift on average. With the success of the pilot, we have now fully-deployed this workflow in Bakken across 200+ gas lift and plunger-assisted gas lift (PAGL) wells. Moreover, the ML-based gas lift optimization workflow presented in this paper is an effective and economic solution for other assets where downhole data or multi-rate testing are not available/feasible due to cost or facility constraints.

## Introduction

For unconventional fields, oil rates typically drop sharply after a few months of production in naturally flowing wells due to the rapid decline of the reservoir pressure. Different kinds of artificial lift methods are then required to continue the production. Gas Lift (GL) is one of the most economic and commonly used artificial lift methods in unconventional fields. During a gas-lift operation, gas is injected into the tubing through a valve from the annulus. The gas mixing with the fluid reduces the mixture density, which decreases the hydrostatic pressure gradient, lowers the flowing bottom hole pressure, increases the pressure draw down and thus ultimately increases the oil flow rate (Brown 1967). When a GL well is in a semi-steady state, its production rate versus gas injection rate typically follows a bell-shaped curve called “Gas Lift Performance Curve” (Figure 1). At a relatively high gas injection rate, the frictional pressure loss starts to compete with the hydrostatic head reduction, resulting in lower production rate. Moreover, the Gas Lift Performance Curve evolves over time as well matures. Therefore, one of the key business questions is to find the optimal gas injection rate that yields the maximum liquid production rate for any given GL well at any time in its life span.

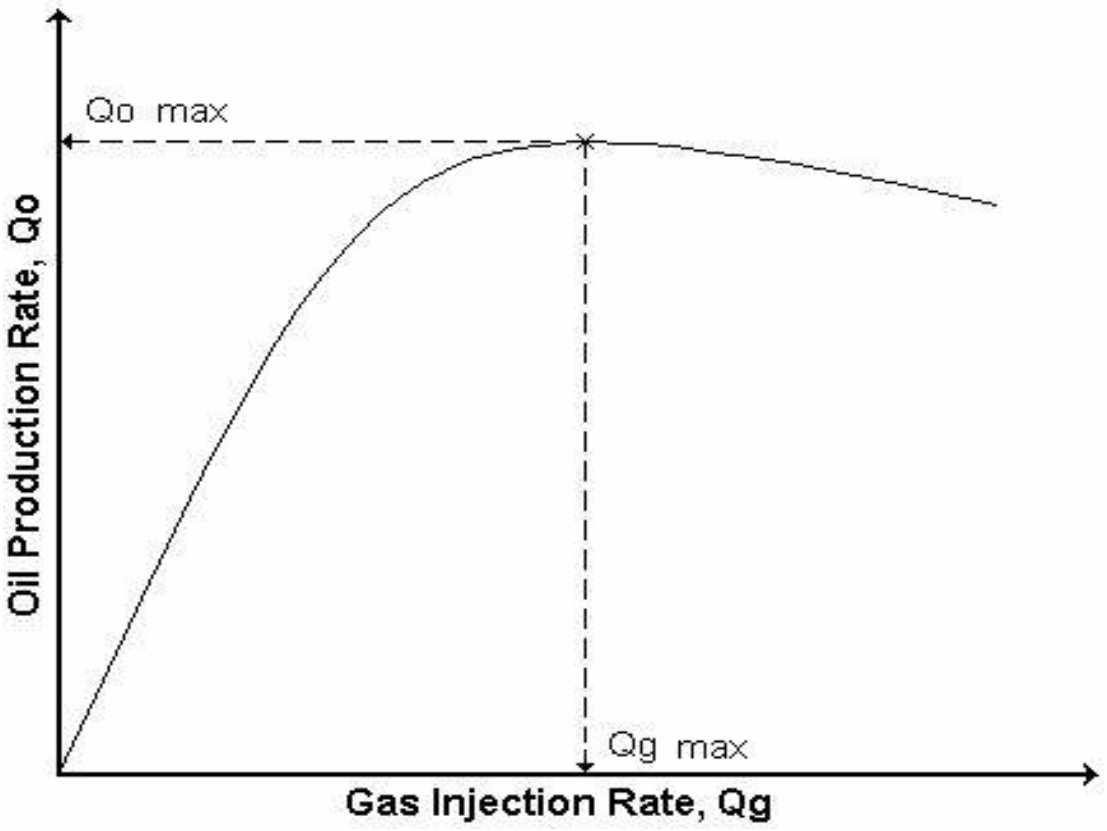

Figure 1. A typical Gas Lift Performance Curve

Physics-based and data-driven approaches are the two typical types of methods solving for Gas Lift Optimization problem (Looi et al., 2023, Borden et al., 2016, Masud et al., 2023, Gambaretto & Rashid 2023, Maut & Prakash, 2023). The former approach usually requires excessive cost/efforts to obtain good estimates of dynamic physical properties and to ensure well-calibrated models throughout wells' lifecycle. Moreover, physics-based approach is usually computationally expensive and may involve assumptions that are not suitable for unconventional fields. These pose challenges to scale up the physics-based methods to hundreds of wells in unconventional fields. Data-driven approach, on the other hand, is easier to deploy and sustain on a full field scale. However, it may require downhole gauges to provide real-time flowing bottom hole pressure data or multi-rate testing to perturb on gas injection rates in a short time window to establish the Gas Lift Performance Curve. Due to the facility/cost constraints, downhole data, or multi-rate well tests may not be available/feasible in many unconventional assets. Thus, many existing data-driven solutions cannot be adopted.

Specifically, due to the corrosive environment and extreme weather conditions, the Bakken asset studied in this paper is sparsely instrumented having no downhole gauges and significant compressor reliability issues. Pad compression is used in Bakken with limited feasible gas changes per week. Furthermore, gas rate adjustment can often create compressor reliability issues, resulting in multiple days of production loss for the whole pad. Therefore, we propose a novel data-driven workflow that does not require downhole data or gas-rate-perturbation tests. We build a machine learning model trained on historical production data to forecast Gas Lift Performance Curves on a weekly basis. Combining this ML forecaster with Bayesian optimization method, the workflow predicts the weekly optimal gas injection rate for every well to maximize the total pad production while not exceeding the compressor capacity. With the optimal gas rate and the potential uplift, the artificial lift team in Bakken can make decisions on when to adjust compressor from risk mitigation perspective in a more scientific and efficient way.

## Theory and Methods

In this section, we detail the workflow using the data from Bakken for illustration. The workflow includes three parts: (1) ML forecaster; (2) Bayesian Optimization and (3) Deployment.

### (1) ML Forecaster

The goal is to build one ML model that learns from the historical production and gas injection rate data to forecast gas lift performance curves for all GL/PAGL wells. For Bakken, we have production rate, gas injection rate, surface pressure data etc. reported daily. Figure 2 shows the raw time series data of some example wells. As can be seen, the liquid production typically has an exponential decay trend over time and the raw daily data is noisy pending clean up. The following data cleaning strategy is applied here.

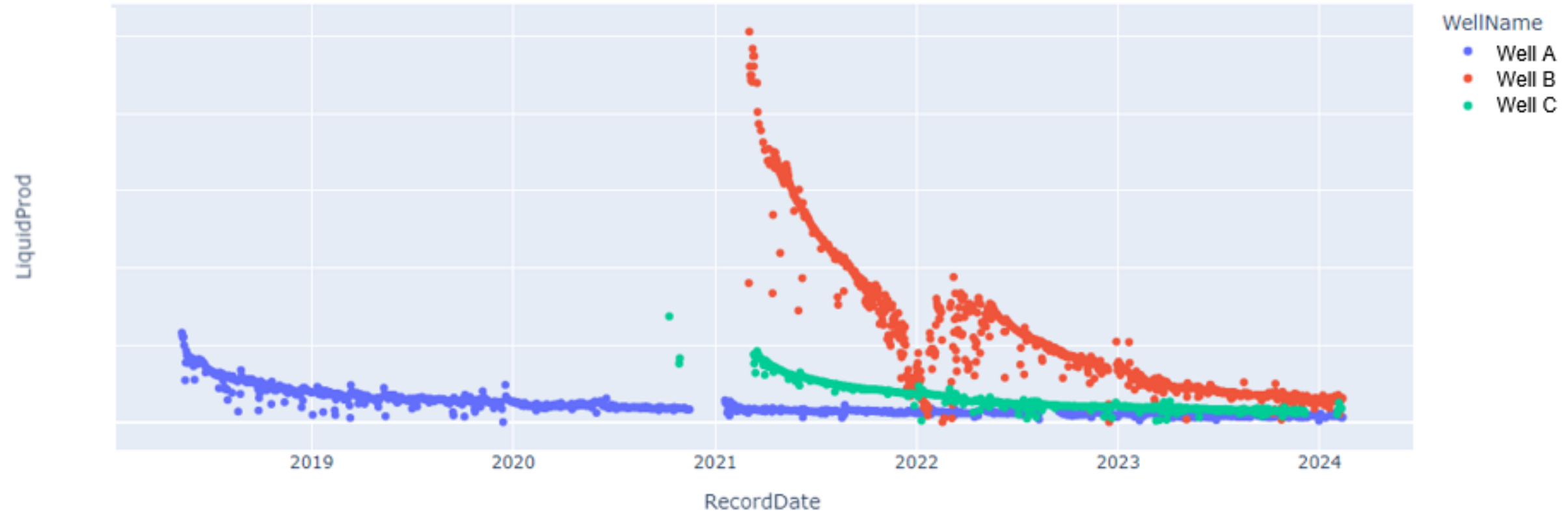

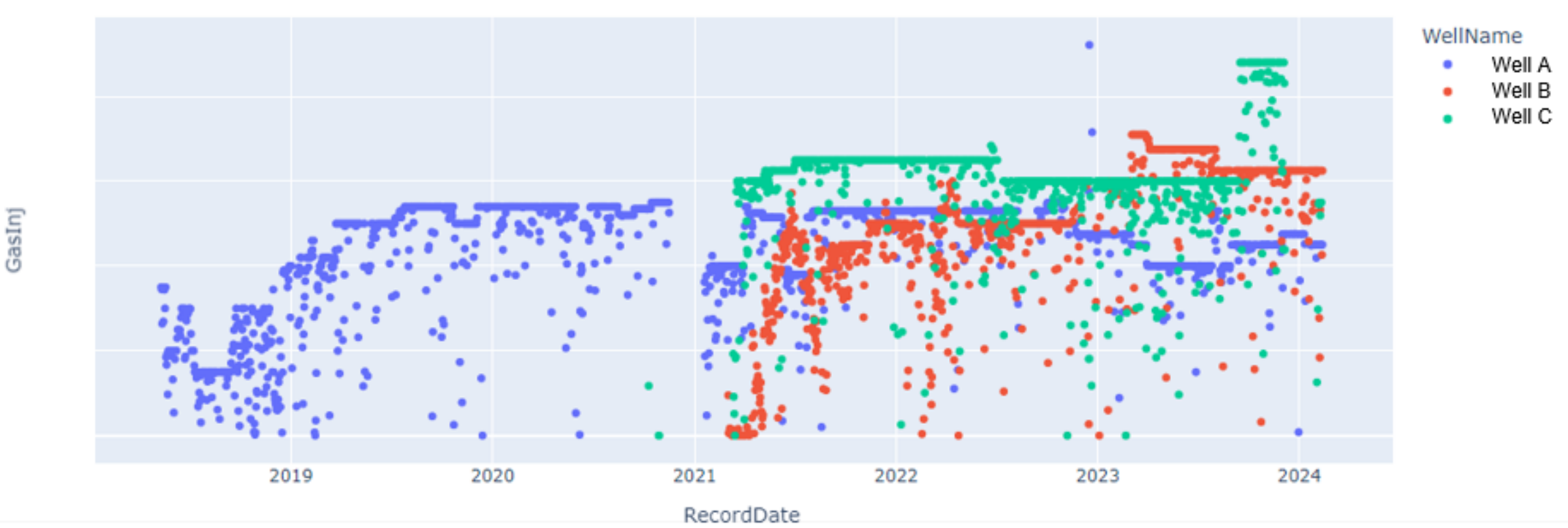

Figure 2. Raw daily time series data for example wells

- Filter Data.
    - Remove first few months of data on GL. Typically the first few months show strong transient effect with gas injection rate slowly ramping up and production rate rapidly dropping down. To prevent biasing the ML model and obtain the Gas Lift Performance Curves at semi-steady state, the very initial few months of data is filtered out.
    - Remove flatlines in daily data.
    - Remove weeks with more than 6 hours downtime in total.
    - Remove days with non-physical production/gas injection data.
    - Remove initial flush out days following long shut-ins.
    - Nullify non-physical surface pressure/temperature time series data.

- Smooth/Down-sample Data.

  We apply median down sampler to aggregate the daily time series data to weekly data to further smooth the data removing outliers. Usually, the optimal gas injection rate would not change dramatically from week to week. In addition, the compressors in Bakken only allow 1-2 gas rate changes per week. Balancing between data smoothness and data size, we choose one week as the down-sample time window.

- Impute Data.

  Forward-filling interpolation scheme is used to impute missing temporal features to prevent data leakage. For static well features (e.g., lateral length, area, total vertical depth…), mean and mode imputations are used for numerical and categorical features, respectively.

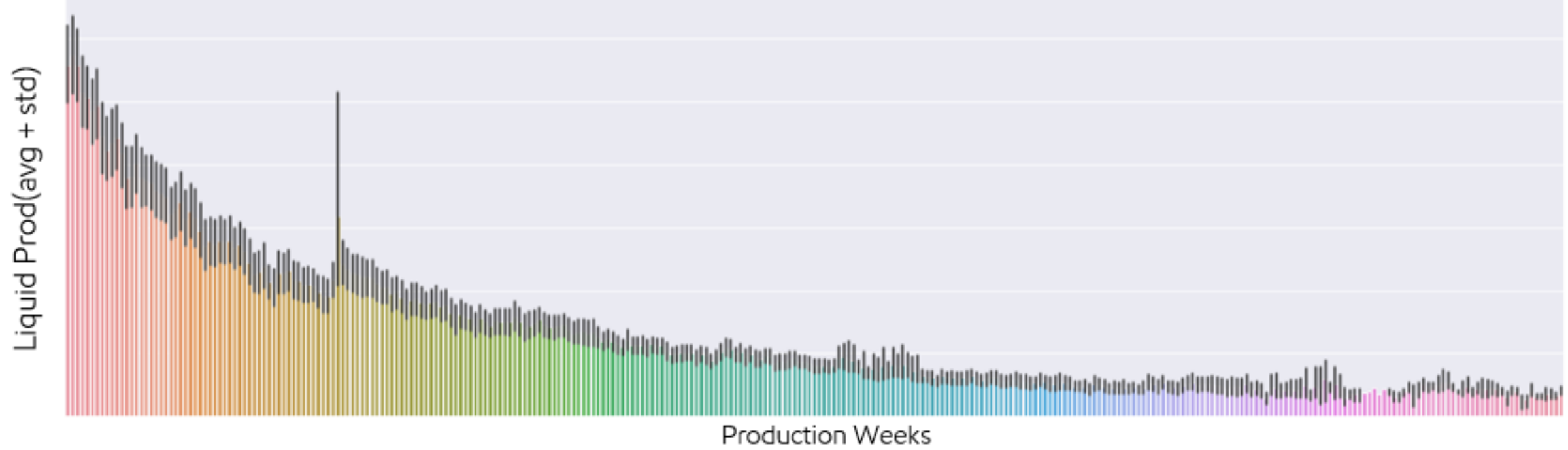

Figure 3. Statistical aggregation of all wells' Liquid Production over production time

After data cleaning, we plot the aggregate of all wells' liquid production over production weeks in Figure 3. Again, we can see the exponential decaying trend in production over time statistically. Like many other time series forecasting problems, the dominant predictor of next week's production should be this week's production. Figure 4 shows the scatterplot of all wells' liquid production at Week(t) versus Week(t-1). As expected, we observe a strong linear correlation between these two. Therefore, we apply the following differencing technique (Equation 1) to transform the time-trending process to a stationary process. With this transformation, we convert this time series forecasting problem into a regression problem with the target $y$ being the nonlinear difference $\Delta LiquidProd(t)$ in Equation 1.

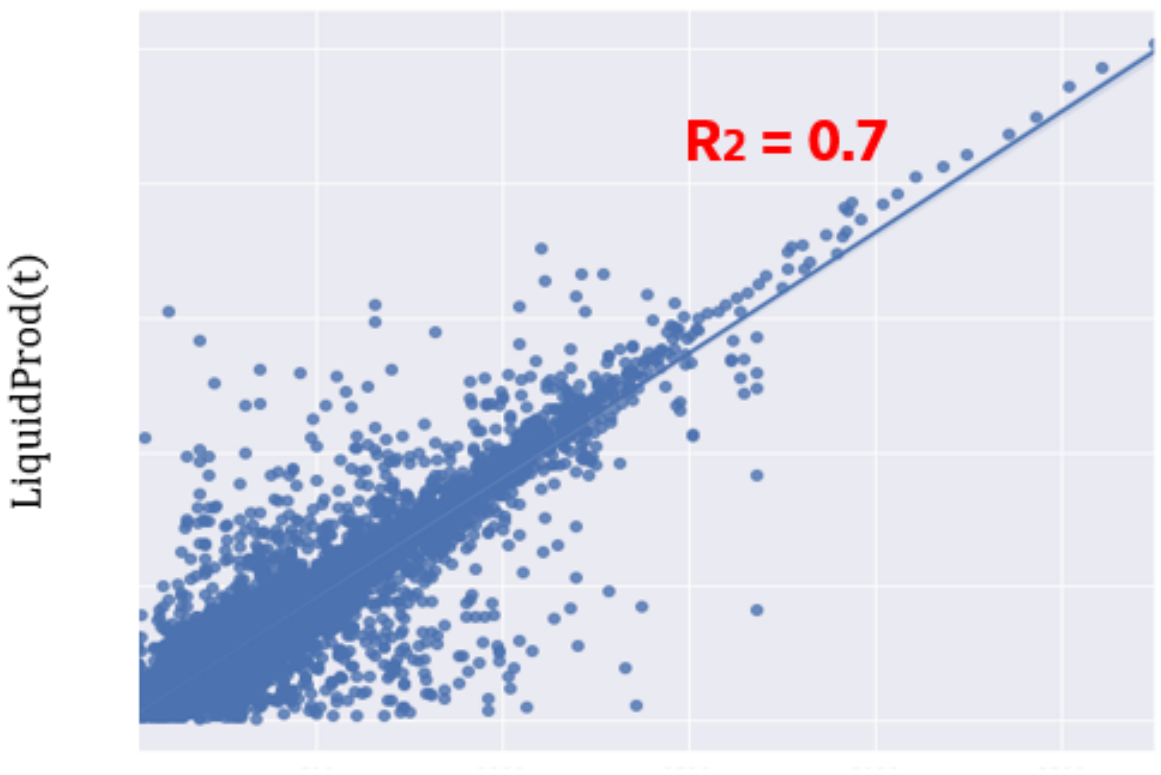

Figure 4. Liquid Production at Week(t) linearly correlates with its value at previous week Week(t-1)

$$LiquidProd(t) = LiquidProd(t-1) - \Delta LiquidProd(t) \tag{1}$$

We formulate the regression problem by mapping three categories of predictors/features to the target $\Delta LiquidProd(t)$ at Week(t), as shown in Figure 5.

- Temporal Features.

  These include rolling/lag time series variables of past liquid production, $\Delta LiquidProd$, surface pressure/temperature at Week(t-3), Week(t-2) and Week(t-1). Note that none of the temporal features should include measurements at Week(t) to prevent data leakage. Here we choose 3 weeks of history for feature engineering as a trade off between capturing recent time trend and utilizing as much data without missing values as possible.

- Control Parameter.

  The control parameter for optimization is the gas injection rate that will be set at the week of prediction Week(t).

- Static Features.

  These include lateral length, total vertical depth, area, production age etc. As for the data normalization scheme, standard

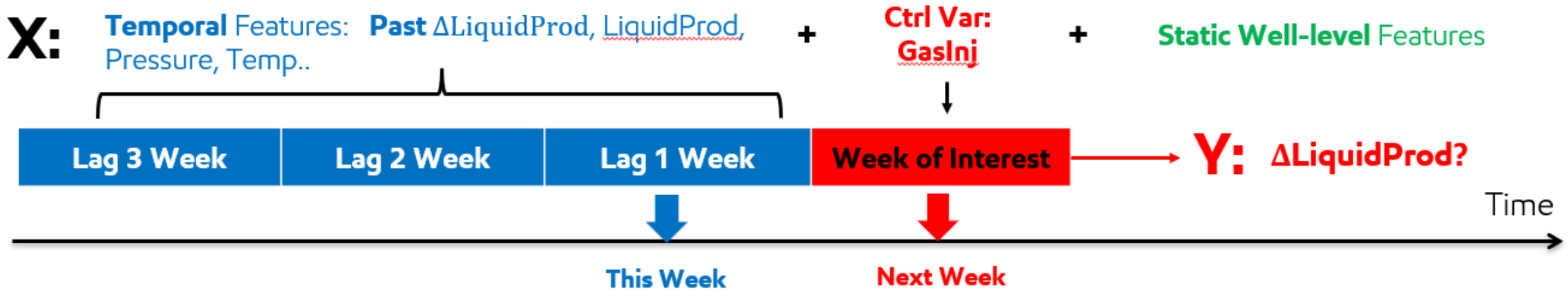

Figure 5. Regression problem formulation with feature engineering

We apply standard scaler to all the numerical features and label encoder to all categorical features from above.

We implemented an ensemble ML model through a simple stacking of a Random Forest (RF; Ho 1995), an XGBoost regressor (XGB; Chen & Guestrin 2016) and a Kernel Ridge Regression model (KRR; Vovk 2013) with a 2$^{nd}$ order polynomial kernel. Through model stacking, we can take advantage of different types of ML models (boosting, bagging and non-tree-based) to further reduce the model variance. Particularly, KRR can improve extrapolation at gas injection rates that are not seen in training data, which pure tree-based models cannot perform well (see Figure 6).

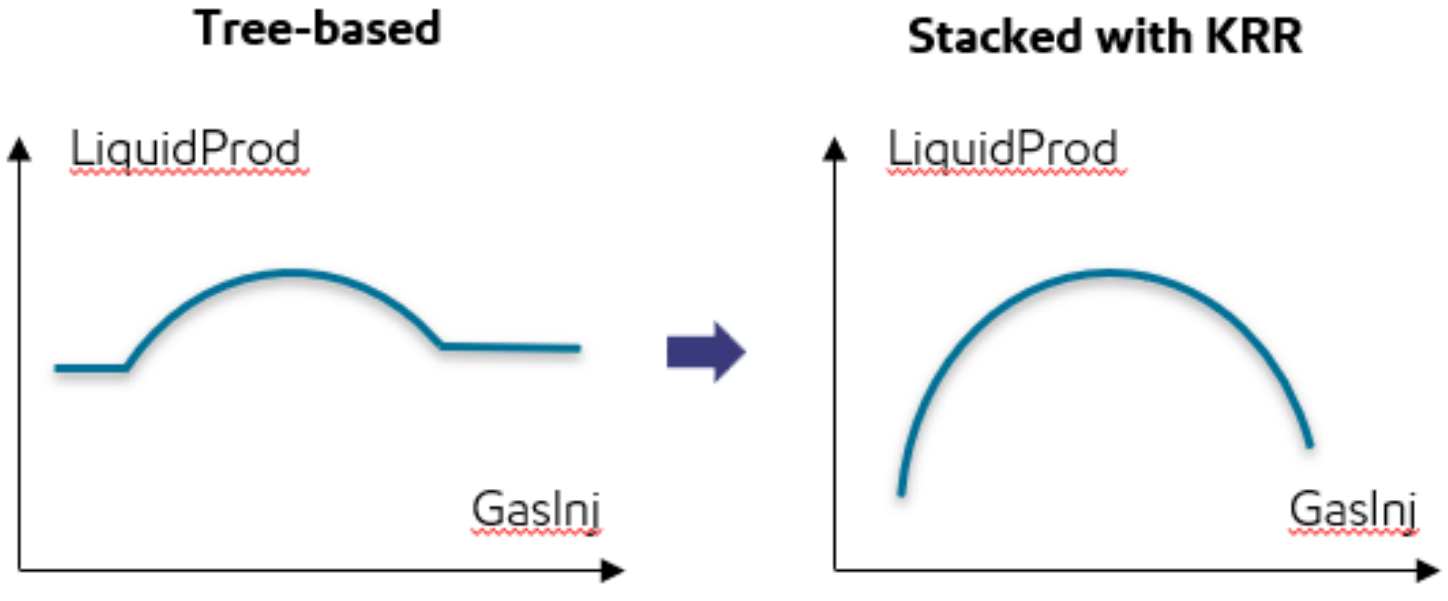

Figure 6. Tree-based models stacked with KRR improves extrapolation.

We tune the ensemble model over a wide range of hyperparameter space using the older 80% (in time) of the data through a time-series cross validation scheme (see Figure 7). This can prevent leaking the past data to the future. Figure 8 shows the prediction results of the final model on both the older 80% training set and the newer 20% hold-out test set. We obtain good accuracy between the predicted and the true liquid production for both training (RMSE = 54.6 bbl/d) and test set (RMSE = 58.bbl/d)

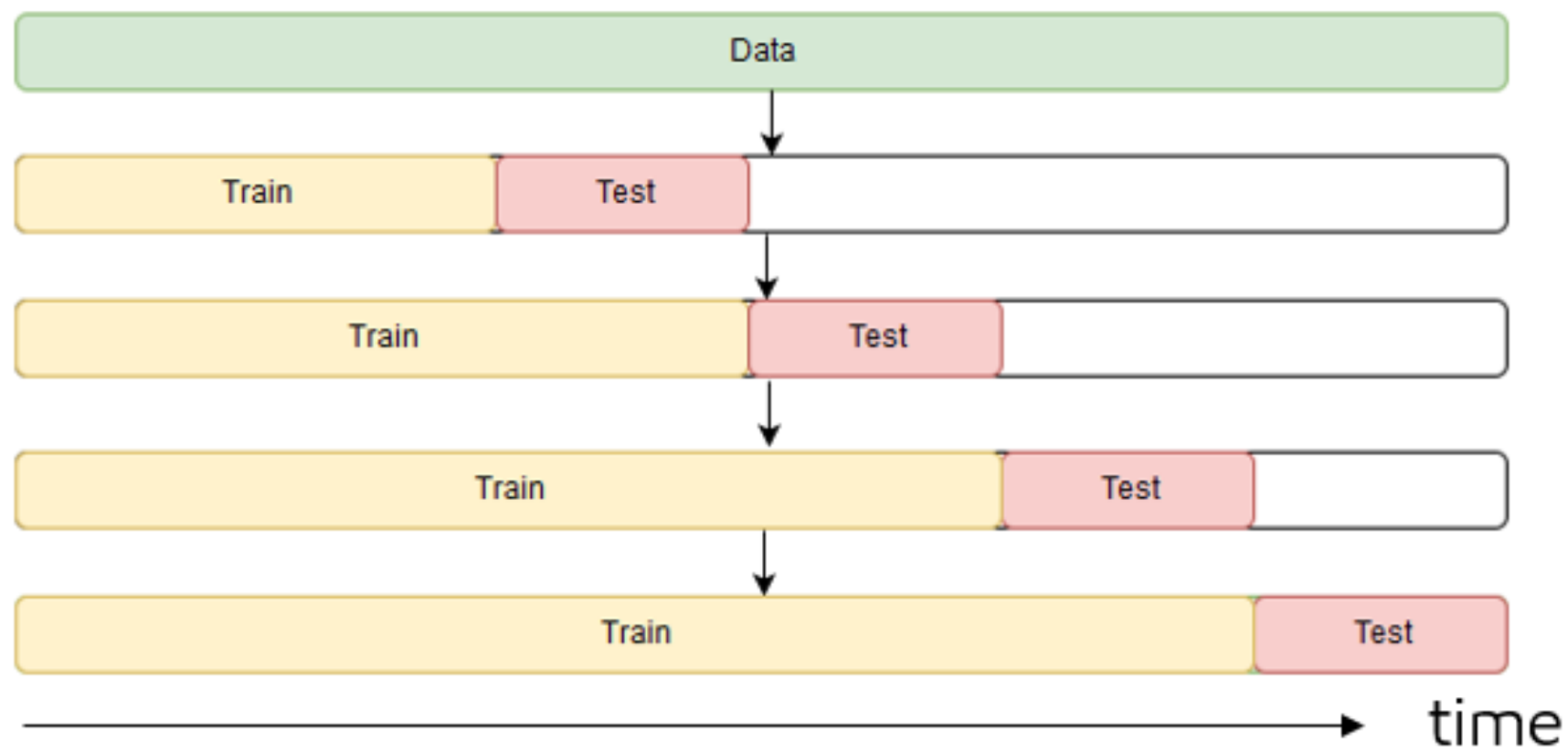

Figure 7. K-fold cross validation scheme for time-series model

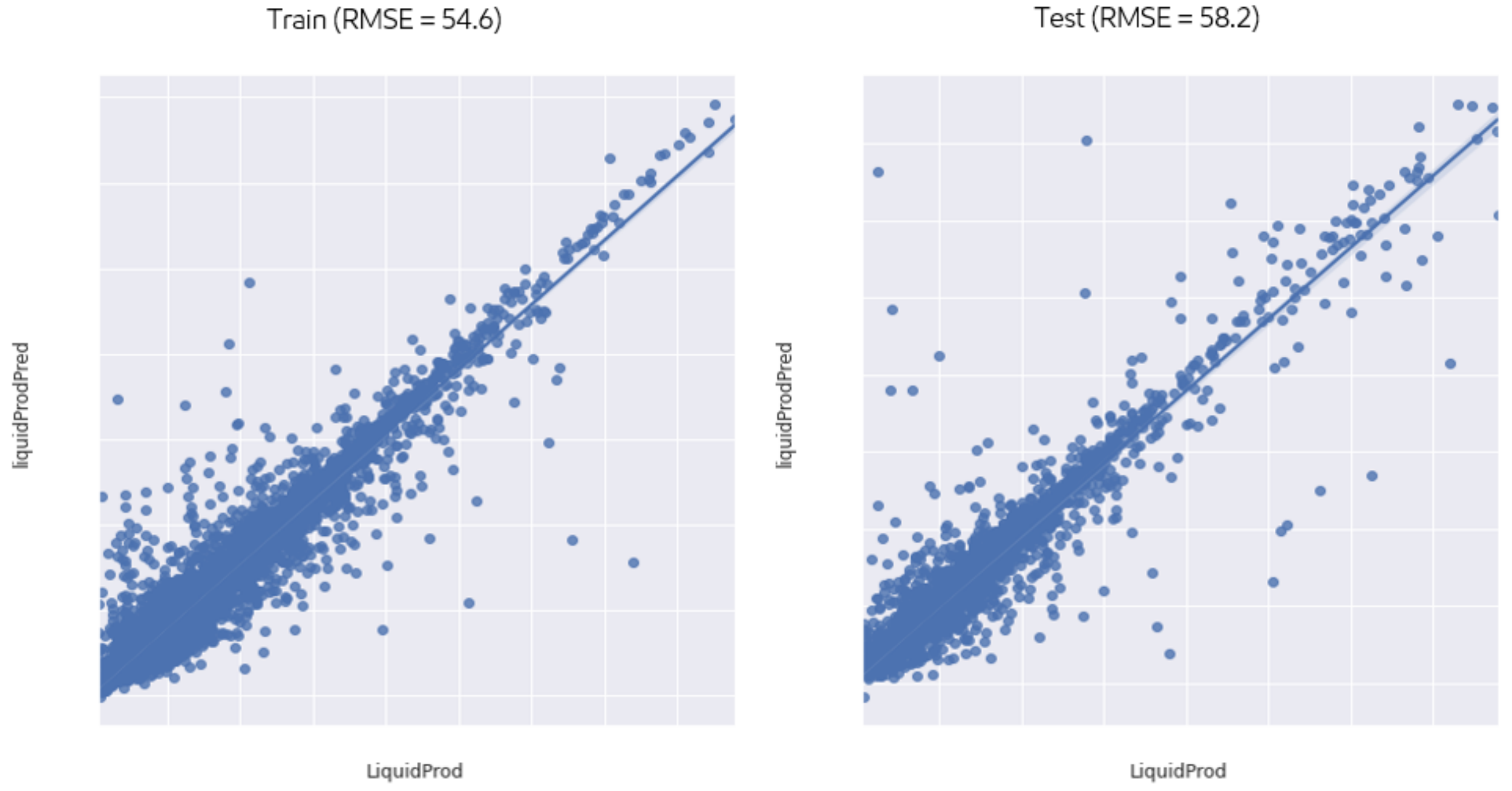

Figure 8. Predicted LiquidProd versus True LiquidProd for training and test set by ML forecaster

The SHAP feature importance from the final ML model is shown in Figure 9. As expected, the most important features are the past liquid production-related temporal features. These features describe the states of the well in terms of the production decline trend. The control parameter GasInj is in the second tier. It proves the importance of production optimization as a secondary effect on top of the dominant decline trend.

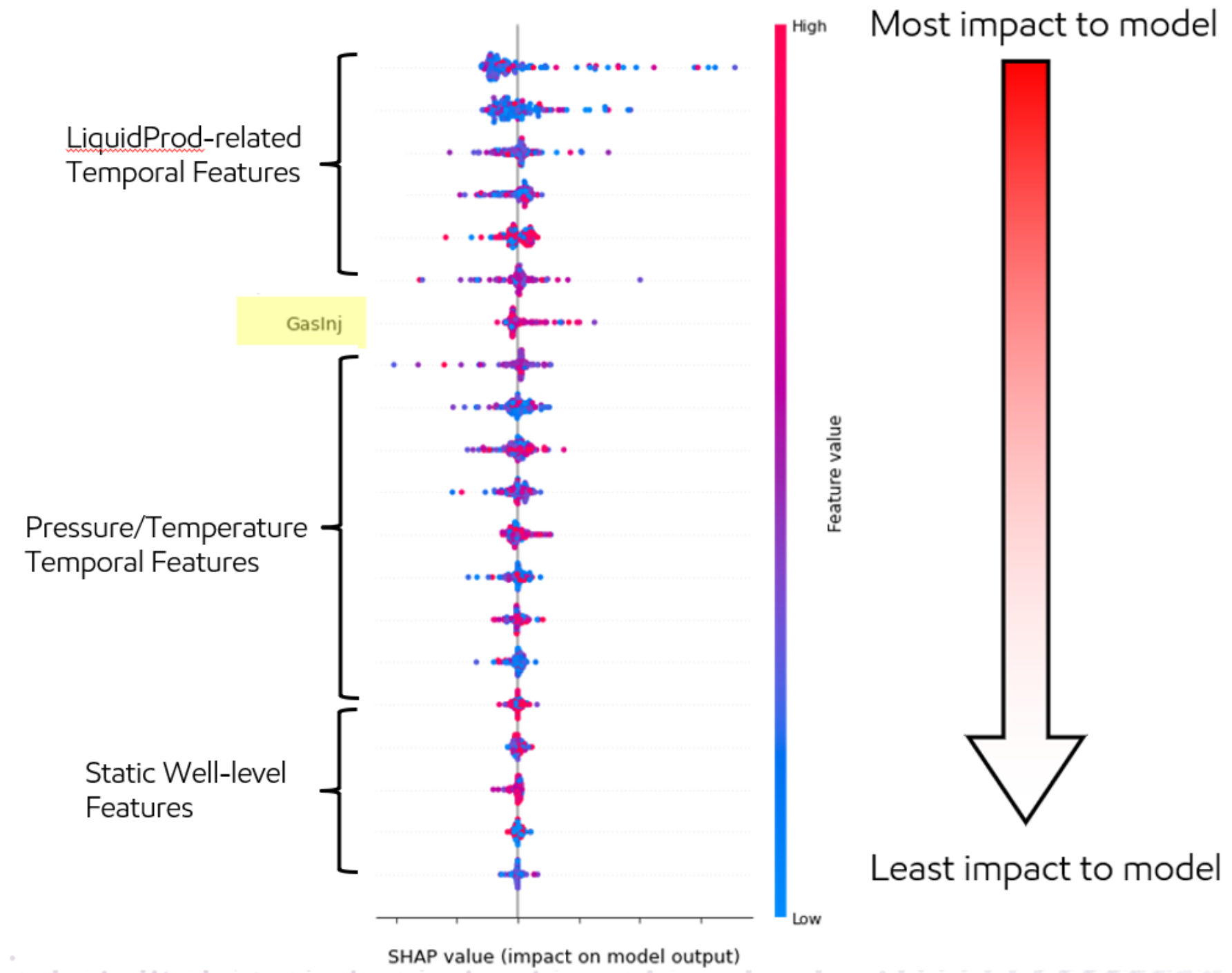

Figure 9. SHAP feature importance of the ML forecaster

Once we have a ML forecaster, we can forecast every well's Gas Lift Performance Curve for the next week by looping through different gas injection rates. We performed sanity check of the ML forecaster to make sure the results are physical following engineers' experience. Typically, we would not expect the Gas Lift Performance Curve varying dramatically from week to week. The unconstrained optimal gas injection rate (corresponding to the maximum production rate in Gas Lift Performance Curve) of a given well should vary on a monthly scale instead of weekly scale. We show two examples in Figure 10. As can be seen, the gas lift performance curves vary slowly over time for both well A and well B. Their corresponding unconstraint optimal gas injection rates change on a monthly scale as expected.

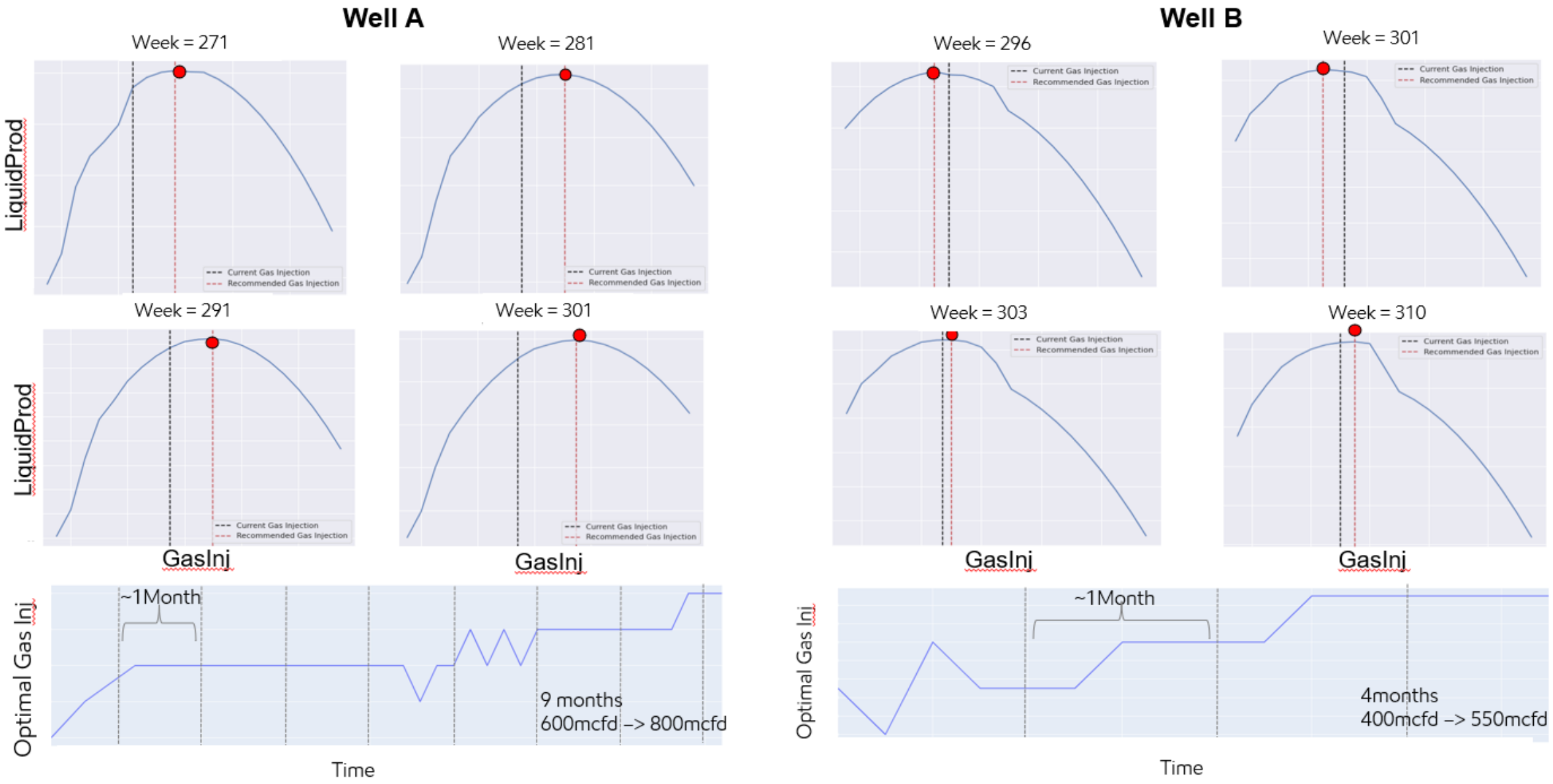

Figure 10 Two examples of Gas Lift Performance Curves varying over time

**(2) Bayesian Optimization**

If there are no facility constraints, the optimal gas injection rate is trivial. It is just the gas injection rate that corresponds to the maximum production rate in the Gas Lift Performance Curve. However, Bakken is on pad compression. Each compressor has its own maximum and minimum required capacity. We further formulate this problem as a constrained optimization problem (see Figure 10).

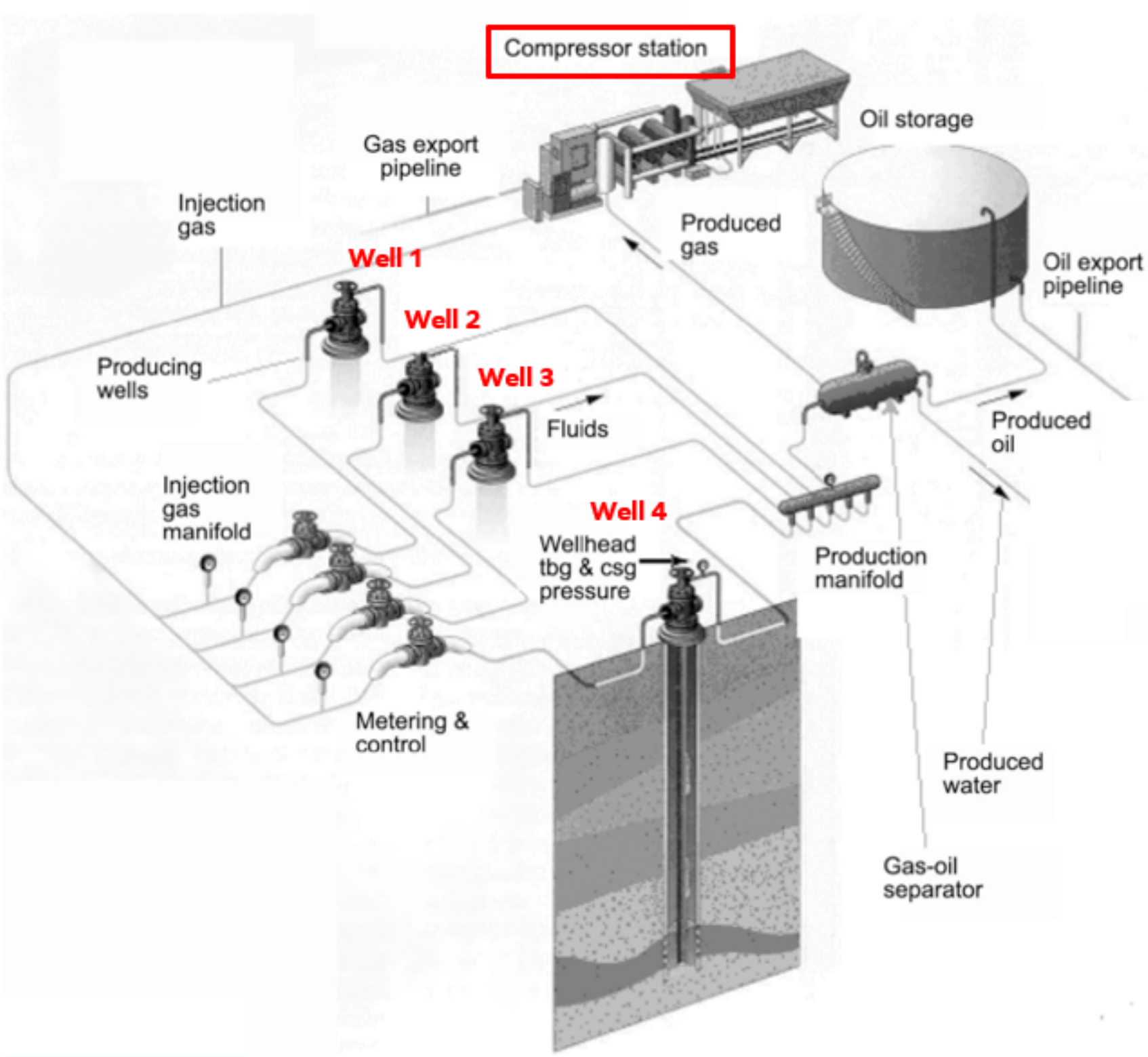

Figure 11. Gas Lift field schematic with wells and compressor

- Given:
  - Wells-to-compressor mapping relationship
  - Compressors' min and max capacities
  - Gas Lift Performance Curves forecasted from ML forecaster
- Find optimal gas injection rate for each well s.t.:
  - Total Production of all wells sharing the same compressor is maximized
  - Total gas injection rates of all wells sharing the same compressor is within compressor's capacity.

Once the constrained optimization problem is formulated, you can choose any optimizer based on your preference. Due to the scale of the problem, we apply a Bayesian optimization method which intelligently explores the parameter space by leveraging past evaluations to guide future selections. This can significantly narrow down the search range and solve the optimization problem in a much more efficient way.

### (3) Deployment

We deployed the above workflow to Bakken through cloud. The underlying architecture consists of two continuous integration continuous deployment (CICD) pipelines (see Figure 10).

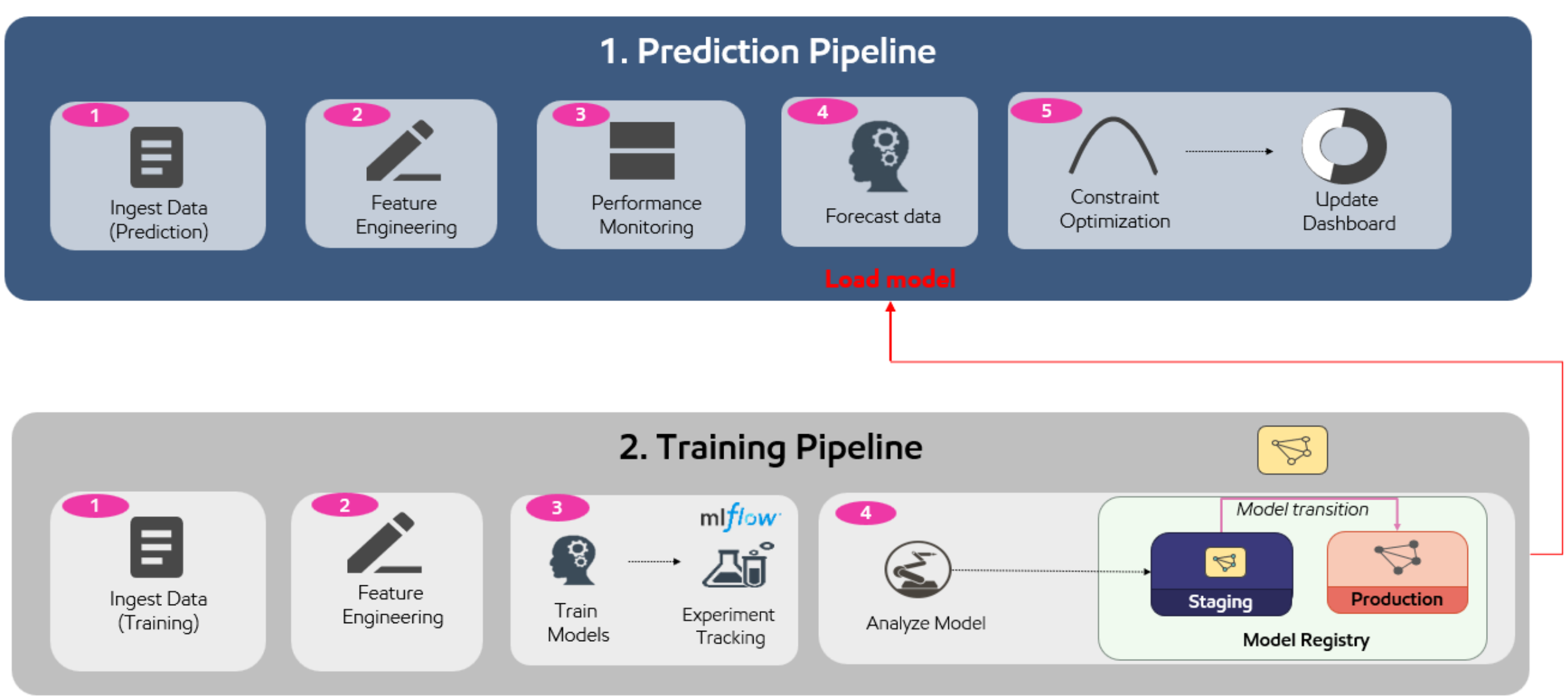

Figure 112. Deployment Architecture of the Gas Lift Optimization Workflow

- Prediction Pipeline

  The prediction pipeline is running on a weekly basis. It serves two purposes. On one hand, it loads a pre-trained ML model to forecast all the wells' Gas Lift Performance Curves and solves for the constrained optimal gas injection rates considering the compressor capacity for the next week. The results are pushed to a dashboard to the end users. On the other hand, it monitors the ML forecaster's performance by comparing last week's prediction against the true measurements. If large deviation has been observed for several weeks in a row, developers will be alerted to do troubleshooting.

- Training Pipeline

  Training pipeline is running monthly. The pre-trained ML model loaded in prediction pipeline is generated through this pipeline. As more wells converted to GL wells and more GL wells converted to PAGL wells, the data distribution would probably shift over time. To capture the data shifting trend, we re-train the ML model every month with the latest data. The new ML model generated will be compared against the existing old model on the same hold-out test set that both models haven't seen. If the new model has better performance (such as RMSE), then it would be automatically promoted. Otherwise, the old model is still up-to-date and can continue running for the next month.

**Results**

Before the full deployment, we did a pilot in Bakken testing this workflow on ~30 wells across 8 different pads. On average we received >5% production uplift. The vertical red dash line Figure 13 shows two examples during the pilot with one well increasing gas injection rate and the other decreasing the gas injection rate. The brown line is the gas injection rate, while the green line is the oil production rate. Both wells show a production increase after the optimization.

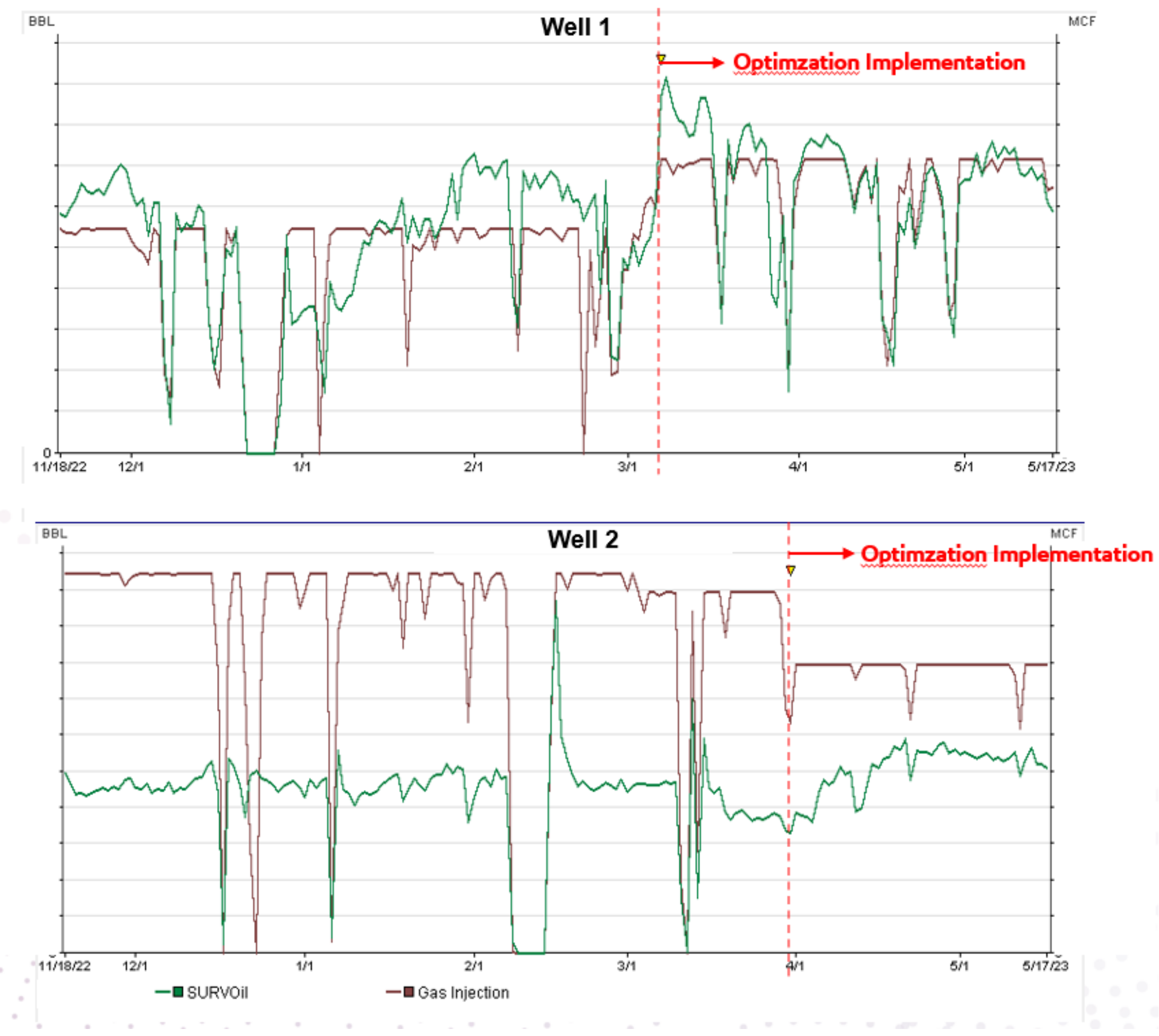

Figure 123. Two examples of Wells Performance using the Optimization Workflow during Pilot

With the success of the pilot, we have now fully-deployed this workflow in Bakken across 200+ GL and PAGL wells. As of today, we already have 50+ wells optimized at least once using this workflow and 84% of the wells are operating in near-optimal state that does not worth the risk of breaking down the unreliable compressors. Among all the wells that have been optimized, we have achieved ~7% production uplift on average. In addition, the web-based dashboard has become the routine tool for the Artificial Lift Team and the Engineering Team during their monthly Surveillance and Optimization meetings in Bakken.

With its help, the engineers can make the decisions on when to worth the risks to adjust the compressor for optimization in a much more efficient and scientific manner. In this way, they can ensure most of the wells are near optimal while adjusting the compressors as little as possible.

**Discussion**

In our case, most of the wells have dedicated separators to record daily production rate. However, this workflow can be easily extended to assets where wells have daily allocated production rate or relatively frequent well tests.

**Conclusions**

We demonstrate an automated data-driven workflow using ML for gas lift optimization in unconventional fields. This workflow has been successfully deployed to Bakken on 200+ GL and PAGL wells since March in 2023. Since deployment, 50+ wells have been at least optimized once using this workflow and >5% averaged production uplift has been obtained. Moreover, this workflow has become Bakken engineers routine surveillance and optimization tool to increase their productivity and efficiency. This ML-based gas lift optimization workflow is easy to scale up and maintain without requiring downhole gauges or multi-gas-rate testing. It is an effective and economic solution for assets that have cost or facility constraints.

**References**

Brown, K. E. 1967. Gas Lift Theory and Practice. Petroleum Publishing Co. United States.

Looi, C.K., Lo, J.H., Husni, M.M. and Abdollahzadeh, A., 2023, October. Enhancing Gas Lift Optimization in Oil Fields: Integrating Data-Driven and Physics-Based Approaches for Improved Decision-Making. In *EAGE Workshop on Data Science-From Fundamentals to Opportunities* (Vol. 2023, No. 1, pp. 1-4). European Association of Geoscientists & Engineers. King, M. J. and Mansfield, M. 1999. Flow Simulation of Geologic Models. SPE Res Eval & Eng 2 (4): 351–367. SPE-57469-PA.

Borden, Z.H., El-Bakry, A. and Xu, P., 2016, September. Workflow Automation for Gas Lift Surveillance and Optimization, Gulf of Mexico. In *SPE Intelligent Energy International Conference and Exhibition*. OnePetro.

Masud, L., Cortez, V.L., Ottulich, M.A., Valderrama, M.P., Ghilardi, J., Sanchez Graff, L.A., Biondi, J. and Sapag, F., 2023, March. Gas Lift Optimization in Unconventional Wells–Vaca Muerta Case Study. In *SPE Argentina Exploration and Production of Unconventional Resources Symposium* (p. D011S004R002). SPE.

Gambaretto, A. and Rashid, K., 2023, October. A Robust Method for Data-Driven Gas-Lift Optimization. In *Abu Dhabi International Petroleum Exhibition and Conference* (p. D031S118R004). SPE.

Maut, P.P. and Prakash, Y., 2023, February. Subsurface Pressure and Temperature Survey Analysis for Wells on Intermittent Gas Lift: A Field Optimisation Case Study in Upper Assam Shelf Basin. In *International Petroleum Technology Conference* (p. D011S002R002). IPTC.

Ho, T.K., 1995, August. Random decision forests. In *Proceedings of 3rd international conference on document analysis and recognition* (Vol. 1, pp. 278-282). IEEE.

Chen, T. and Guestrin, C., 2016, August. Xgboost: A scalable tree boosting system. In *Proceedings of the 22nd acm sigkdd international conference on knowledge discovery and data mining* (pp. 785-794).

Vovk, V., 2013. Kernel ridge regression. In *Empirical Inference: Festschrift in Honor of Vladimir N. Vapnik* (pp. 105-116). Berlin, Heidelberg: Springer Berlin Heidelberg.